\documentclass{article}

\usepackage[utf8]{inputenc}
\usepackage[style=numeric]{biblatex}
\usepackage[preprint, nonatbib]{neurips_2024}
\usepackage{booktabs}

\usepackage{bm}
\usepackage{amsmath}
\usepackage{svg}
\usepackage{float}
\usepackage[utf8]{inputenc} 
\usepackage[T1]{fontenc}    
\usepackage{hyperref}       
\usepackage{url}            
\usepackage{booktabs}       
\usepackage{amsfonts}       
\usepackage{nicefrac}       
\usepackage{microtype}      
\usepackage{xcolor}         

\title{Adaptive Variational Continual Learning via Task-Heuristic Modelling}

\author{%
  Fan Yang \\
  Department of Computer Science\\
  University of Oxford\\
  Oxford, UK.\\
  \texttt{fan.yang@cs.ox.ac.uk} \\
}

\begin{document}

\maketitle

\begin{abstract}
   Variational continual learning (VCL) is a turn-key learning algorithm that has state-of-the-art performance among the best continual learning models. In our work, we explore an extension of the generalized variational continual learning (GVCL) model, named AutoVCL, which combines task heuristics for informed learning and model optimization. We demonstrate that our model outperforms the standard GVCL with fixed hyperparameters, benefiting from the automatic adjustment of the hyperparameter based on the difficulty and similarity of the incoming task compared to the previous tasks.
\end{abstract}

\section{Introduction}

Continual learning represents a trending topic in AI research, aiming to create systems capable of incrementally acquiring new skills while retaining previously learned knowledge\cite{chen2018lifelong, hadsell2020embracing}. This objective mirrors the human intelligence for lifelong learning but consists of notable challenges such as catastrophic forgetting\cite{mccloskey1989catastrophic}, where newly acquired knowledge can disrupt what has been previously learned. The aim for scalable and efficient solutions is crucial, as the integration of new data often leads to increased model sizes and computational demands\cite{parisi2019continual}. Additionally, the development of strategies that maintain a balance between preserving old knowledge and incorporating new insights is equally important\cite{abraham2005memory}. Despite these obstacles, progress in continual learning is fundamental for advancing AI systems that are adaptable and intelligent, capable of adjusting to their environments and experiences\cite{ring1994continual, khetarpal2022towards}.

\paragraph{Variational Continual Learning (VCL)}represents a milestone advancement in the development of artificial intelligence systems capable of learning continuously over time \cite{nguyen2018variational}. This method leverages Bayes' rule to compute a posterior distribution over the model parameters \(\bm{\theta}\), utilizing a prior parameter distribution \(p(\bm{\theta})\) for a specific dataset \(\mathcal{D}_t\). Derived from Bayes' theorem, this method naturally facilitates online and continual learning processes through variational inference of the posterior, using the preceding posterior as the new prior combined with the likelihood of new data. In their study, \textcite{nguyen2018variational} demonstrated that recursively updating the posterior distributions of \(\bm{\theta}\) captures the predictive capability accumulated from observing all data up to the current point. Furthermore, the authors addressed the computational challenges associated with having to analytically solve intractable terms by approximating the posterior \(p(\bm{\theta})\) with a more manageable distribution \(q_t(\bm{\theta})\), typically Gaussian, using the Kullback–Leibler (KL) divergence.

The optimal approximated distribution is found using the evidence lower bound (ELBO), which is defined as 
\[\text{ELBO} = \mathbb{E}_{\bm{\theta} \sim q_t(\bm{\theta})}[\log p(\mathcal{D}_{t} |\bm{\theta})] - \text{KL}(q_t(\bm{\theta}) \| q_{t-1}(\bm{\theta})),\]

for \(t\in\{1,2,\dots,T\}.\)
The first term, \(\mathbb{E}_{\bm{\theta} \sim q_t(\bm{\theta})}[\log p(\mathcal{D}_{t} |\bm{\theta})]\), is the log-likelihood of observing the data given our existing model parameters, which represents how well our current model explains the data. The KL divergence in the second term measures how much our approximated posterior deviates from the last state, which conceptually preserves the existing knowledge of previously learned tasks. 

Building upon the framework of variational inference, \textcite{nguyen2018variational} enhanced their training algorithm by incorporating an episodic memory structure, namely the \textbf{coreset}. This strategy involves selecting and retaining a subset of data specific to each task. When introduced to new tasks, the model refreshes its memory by revisiting the coreset, thereby enhancing the performance of VCL with minimal additional training resources. Moreover,  \textcite{nguyen2018variational} demonstrated that the coreset alone does not significantly influence the prediction accuracy, which remarks the importance of applying the VCL algorithm. 

As shown in the split MNIST experiment from the original study \cite{nguyen2018variational} (also reproduced \ref{appendix}{here}), the accuracy trends of different tasks vary significantly over sequential training. For instance, the accuracy for the 0/1 classification remains nearly perfect after learning five tasks, whereas the accuracy for differentiating \label{drop}between 2/3 significantly decreases from 98\% to 92\% after the fifth task. This variation in task performance is also highlighted in their study, where "adversarial ordering" is suggested as a direction for future research\cite{nguyen2018variational}. In our study, we test the robustness of the original VCL framework beyond the experiments used by \textcite{nguyen2018variational}. Furthermore, we introduce an augmented VCL architecture that adapts to different task sequences, resulting in superior performance compared to the original VCL, in both standard and adversarial task orderings.
\section{Related work}

\paragraph{Continual Learning by modelling Intra-Class Variation (\textcite{modellingtask})}  Efforts have been made to model the difference between tasks in order to mitigate problems in continual learning. \textcite{modellingtask} presents a framework named MOCA (modelling Intra-Class Variation for Continual Learning with Augmentation), designed to alleviate catastrophic forgetting by enhancing representation diversity through two types of perturbations: model-agnostic and model-based. The model-agnostic approach in MOCA diversifies representations using isotropic Gaussian distributions, while the model-based approach incorporates Dropout-based augmentation (DOA), weight-adversarial perturbation (WAP), and variation transfer (VT). These techniques leverage the model's parameters or the features of new classes to create informative perturbations. The paper presents the feasibility and effectiveness of modelling intra-class diversity; in our study, we want to capture instead of perturbation for improved learning outcomes.

\paragraph{Generalized Variational Continual Learning (\textcite{loo2020generalized})} 
Beyond the foundational study of variational continual learning, \textcite{loo2020generalized} proposed a generalized variational continual learning (GVCL) framework that leveraged a flexible $\beta$-ELBO goal for optimization. The $\beta$-ELBO is defined
\[\textcolor{blue}\beta\text{-ELBO} = \mathbb{E}_{\bm{\theta} \sim q_t(\bm{\theta})}[\log p(\mathcal{D}_{t} |\bm{\theta})] - \textcolor{blue}{\beta}\text{KL}(q_t(\bm{\theta}) \| q_{t-1}(\bm{\theta})),\]
where the vanilla VCL is trivially recovered when $\beta = 1$. The magnitude of $\beta$ effectively controls the focus of optimization upon a specific task, where we seek to use this formula to adaptively frame our model for each specific task sequence.
\section{Proposed approach}
\label{approach}

The proposed method aims to tackle the challenges posed by task differences, such as the accuracy decline observed in the Split MNIST example, by adapting the $\beta$-ELBO objective from the GVCL study to develop an adaptive continual learning strategy for different tasks. Ideally, optimization in relation to the ELBO seeks to find a balance between minimizing the model's reconstruction error and reducing deviation from the existing model. Instead of maintaining a fixed weight ratio between these two objectives, our goal is to discover a balance formula that adjusts to different learning tasks, thereby achieving optimal and robust training performance for all sequences of tasks. The detailed modelling approach is outlined in Section \ref{theory}.

For the experimentation, we utilized the proposed multi-head discriminative framework as our model architecture, adding task-specific ``head networks'' after a foundational network with parameters that are shared across tasks. During back-propagation, the shared parameters are updated according to the $\beta$-ELBO, while the task-specific layers are optimized only during training on their respective tasks.

\section{Theory}
\label{theory}
In the optimization algorithm, a small $\beta$ prioritizes minimizing the reconstruction error, while a large $\beta$ focuses more on minimizing the KL divergence between the approximated posterior and the current prior. The appropriate magnitude of $\beta$ is chosen considering two metrics: the \textbf{difficulty} relation between the new task and old tasks, and the \textbf{similarity} between the learned knowledge and the new task. 

For the difficulty metric, we classify the training scenarios into three categories to adjust our $\beta$ value for optimal training performance. When the new task is less difficult compared to the previous tasks, a higher $\beta$ value is preferred to preserve knowledge from previous data. When the new task is more difficult than the previous tasks, the constraint on keeping the posterior close to the prior should be relaxed. Consequently, the model shifts its focus towards the new tasks, with an increased risk of forgetting. When the tasks are of comparable difficulty, $\beta=1$ is used to balance the reconstruction loss and KL divergence. Regarding the similarity metric, a larger $\beta$ should be chosen if the new task closely resembles previous tasks and vice versa. This approach aims to preserve the knowledge already acquired, as it should be transferable to new tasks.

\subsection{Quantification of the metrics}

To quantify the concepts of difficulty and similarity, we employ the existing model and a limited random selection of new training data. The difficulty of a new task \(d_t\) is defined as \begin{equation*}d_t = \min\left(\max\left(\frac{a_t - a'_t}{1-a'_t}, 0 \right),1\right),
\end{equation*} where \(a_t\) represents the accuracy obtained from one-epoch training of a small data subset on an untrained model with the same architecture, and \(a'_t=1/\)(\# of outcomes) denotes the baseline accuracy achievable through random guesses. This formulation of the difficulty metric aims to evaluate how much improvement a mock training would accomplish using limited data and compute.

The similarity of a new task \(t\) is defined as \begin{equation*}s_t =\text{norm}(|a^*_t - a'_t|; 0, 1-a'_t)
\end{equation*} where \(a^*_t\) is the accuracy of the current model's raw predictions on the new task before additional training. The norm function acts as a sigmoid-like function that projects \(|a^*_t - a'_t|\) to the range \([0,1]\), where norm\((|a^*_t - a'_t|) \approx 0\) when \(|a^*_t - a'_t| =0\) and norm\((|a^*_t - a'_t|) \approx 1\) when \(|a^*_t - a'_t| =1- a'_t.\) The similarity metric evaluates the performance of the existing model to the new task before training.

\subsection{Formulation of $\beta_t$ }

For an iterative training sequence, we define $$\beta_t =\exp\left(\lambda\bigg(\max(\{d_1,d_2,\dots,d_{t-1}\}) - \frac{d_t}{1+\delta_d(t-1)}+s_t\bigg)\right),$$ where $\delta_d$ represents the average difficulty gap observed in the previous tasks. This formula follows the guidelines of Section \ref{theory}, incorporating a factor of $\lambda$ to adjust the \(\beta\) magnitude to a practical range, thus ensuring that variations in \(\beta\) values are significant and impactful. The scaling factor $\frac{1}{1+\delta_d(t-1)}$ for the current task's difficulty aims to optimize average accuracy effectively. If tasks of varying difficulty levels have been previously learned, $\beta_t$ should remain high to preserve the posterior, even at the cost of the next single task's accuracy. 

\section{Experiments}

We selected three sequences of tasks that reflect distinct scenarios for continual learning. For each sequence, our AutoVCL model was compared to the GVCL using constant $\beta$ values of 0.01, 1, and 100. We selected $\lambda=5$ as the scaling factor for $\beta_t$. All experiments were conducted without coreset replay, and the data represents the average of five experimental trials. The network architecture chosen for each experiment resembles the original setup in the VCL work for comparison\cite{nguyen2018variational}.

\paragraph{Split MNIST with Custom Targets:} The Split MNIST dataset is a widely used dataset for binary classifications\cite{nguyen2018variational, zenke2017continual}. We aimed to create a sequence of tasks that were intentionally made similar, arranging the sequence as 0/1, 8/7, 9/4, 6/2, 3/5. We employed multi-head networks with two hidden layers, each containing 256 units with ReLU activations.

\paragraph{Permuted MNIST:} The Permuted MNIST dataset serves as a popular dataset in continual learning and was also applied in the VCL work\cite{nguyen2018variational, goodfellow2013empirical}. Each task in the sequence involves images transformed by a unique random permutation, presenting a classic challenge of tasks with similar difficulties but no similarities. For all models, we utilized fully connected single-head networks with two hidden layers, each layer comprising 100 hidden units and ReLU activations.

\paragraph{Mixed Dataset of Split-CIFAR-10 and Split-MNIST} CIFAR-10 is known to be more challenging for prediction than MNIST\cite{krizhevsky2010convolutional}. In this set of experiments, we adhered to the standard configuration with 0/1, 2/3, 4/5, 6/7, 8/9 splits for both MNIST and CIFAR-10. We sequenced the binary classification tasks in an alternating manner between MNIST and CIFAR-10, simulating a scenario in continual learning where the model is exposed to tasks of varying difficulties. For all models, multi-head networks with two hidden layers, each comprising 256 units with ReLU activations were used.
\begin{figure}
  \centering
  \includegraphics[width=\textwidth]{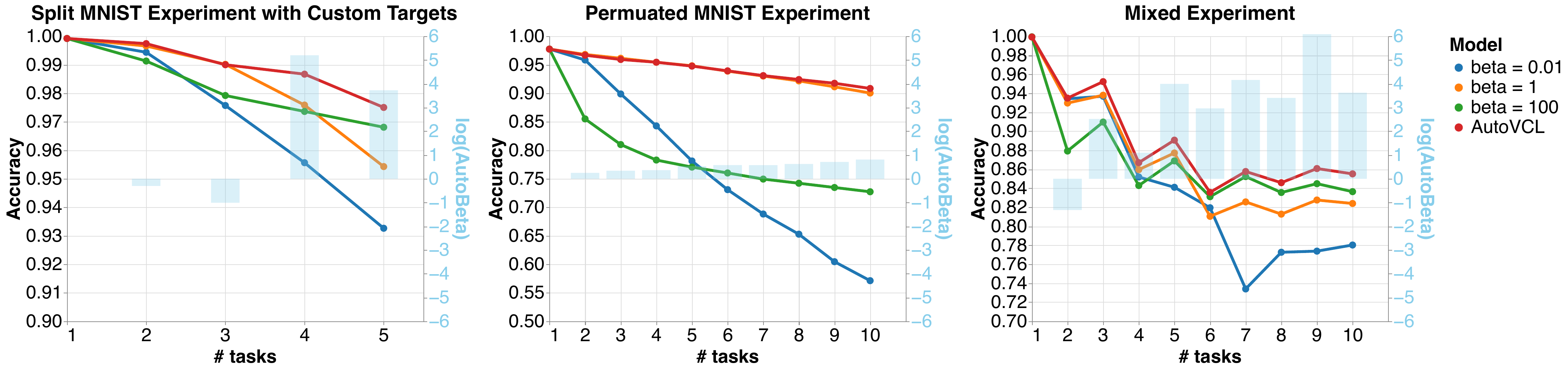}
  \caption{Comparison between AutoVCL and fixed $\beta$ values in three experiments. \textbf{Split MNIST with Custom Targets}, \textbf{Permuted MNIST}, and \textbf{Mixed Dataset of Split-CIFAR-10 and Split-MNIST} respectively represent sequences of similar tasks, different tasks, and alternating difficulty tasks.}
  
\label{res}
\end{figure}

Figure \ref{res} shows the performance of GVCL with fixed $\beta$ values during training as well as the performance comparison of the proposed AutoVCL model. The second $y$-axis illustrates the calculated $\beta_t$ on a logarithmic scale. When $\beta = 1$ (or $\log(\beta_t)=0)$, we replicate the original VCL model proposed by \textcite{nguyen2018variational}. In the split MNIST experiment with custom targets, AutoVCL shows performance comparable to the $\beta=1$ GVCL model in the early tasks. As training progresses, $\beta_t$ is adjusted higher, similar to the $\beta = 100$ GVCL model, achieving over 97\% accuracy, whereas the $\beta=1$ GVCL model's accuracy drops to 95\% after learning the fifth task. In the permuted MNIST experiment, AutoVCL and $\beta = 1$ GVCL achieve 91\% and 90\% accuracy respectively after 10 tasks, while $\beta = 0.01$ GVCL and $\beta = 100$ GVCL achieve 58\% and 73\% respectively. In the mixed experiment involving CIFAR-10 and Split MNIST, AutoVCL matches or exceeds the performance of low $\beta$ GVCL models and maintains a stable trend in the later stages of learning, similar to the $\beta = 100$ GVCL. It achieves 86\% accuracy after 10 tasks, while $\beta =0.01,$ $\beta = 1,$ and $\beta = 100$ GVCL models achieve 72\%, 82\%, and 84\% accuracy respectively.

\section{Discussion}

The experiments described in the previous sections highlight the significance of adjusting $\beta$ magnitude along the training process. When $\beta$ approaches 0, we revert to the maximum likelihood estimation. In the mixed experiment, a model operating under a small constant $\beta$ can outperform other models during the initial stages because of its capability of quickly learning new tasks. However, the model with a small beta tends to experience catastrophic forgetting over the long term. Conversely, a model with a large fixed $\beta$ struggles to match the accuracies of other models initially but can maintain consistent performance over time. Notably, it is observed that in various experimental setups, no single fixed $\beta$ outperforms other GVCL settings in all circumstances.

Our proposed AutoVCL model leverages the benefits of both large and small magnitudes of $\beta$. When the upcoming task is difficult, the model relaxes the constraint on the KL divergence to facilitate better learning of the new data. Conversely, when the next task is relatively easy or can be predicted using the existing model to some extent, the model applies a large $\beta$ focusing on preservation of the knowledge from previous tasks, as learning the new tasks does not require significant changes in the model parameters. As demonstrated by the three experiments, not only does AutoVCL surpass the performance of other models after 10 learning tasks, but it also proves to be the optimal or near-optimal algorithm along each stage of the training process.

In continual learning, the complexity and transferability of future tasks are often unforeseen. AutoVCL offers a method to automatically quantify and adjust for an improved training strategy upon encountering a new task. By taking into account both prediction accuracy and parameter distribution shift, AutoVCL is designed to minimize unnecessary changes to the model in order to maintain equivalent performance; at the same time, it seeks to maximize model accuracy when a shift in parameter distribution is needed.

Being able to leverage task heuristics empowers us to design more flexible systems. In particular, models are likely to benefit from the additional support of coresets. A future research direction could explore how to utilize the properties of tasks for informed training, such as replay scheduling based on task heuristics\cite{schedule}. For example, during coreset replay, more challenging tasks could be re-trained first with a small $\beta$ to achieve higher prediction accuracy, while easier tasks could follow, trained with a high $\beta$ while loosening the KL constraint only on the task-specific layer. Additionally, a notable limitation of our AutoVCL approach is the requirement for repeated trials to obtain a relatively stable assessment of task difficulty. This requirements introduces additional computational complexity. In the future, the exploration of more effective metrics for task evaluation could lead to better outcomes.

\section{Conclusion}

In conclusion, we present AutoVCL, an enhancement of the variational continual learning model, by incorporating a self-adjusted $\beta$-ELBO loss function. We captured task variations through quantification metrics of difficulty and similarity, which the learning algorithm utilizes for informed ratio adjustment of reconstruction error and KL divergence balance. This approach yields optimal performance not only in terms of final averaged accuracy but also throughout the entire learning trajectory. Notably, AutoVCL excels across a diverse range of tasks without the necessity for hyper-parameter tuning, in contrast to $\beta$-GVCL, in which no single $\beta$ is suitable for every type of task sequence. This work provides an exemplar groundwork for future research into more effective task heuristics modelling.



\printbibliography

@misc{nguyen2018variational,
      title={Variational Continual Learning}, 
      author={Cuong V. Nguyen and Yingzhen Li and Thang D. Bui and Richard E. Turner},
      year={2018},
      eprint={1710.10628},
      archivePrefix={arXiv},
      primaryClass={stat.ML}
}

@article{hadsell2020embracing,
  title={Embracing change: Continual learning in deep neural networks},
  author={Hadsell, Raia and Rao, Dushyant and Rusu, Andrei A and Pascanu, Razvan},
  journal={Trends in cognitive sciences},
  volume={24},
  number={12},
  pages={1028--1040},
  year={2020},
  publisher={Elsevier}
}

@article{krizhevsky2010convolutional,
  title={Convolutional deep belief networks on cifar-10},
  author={Krizhevsky, Alex and Hinton, Geoff},
  journal={Unpublished manuscript},
  volume={40},
  number={7},
  pages={1--9},
  year={2010},
  publisher={Citeseer}
}

@article{goodfellow2013empirical,
  title={An empirical investigation of catastrophic forgetting in gradient-based neural networks},
  author={Goodfellow, Ian J and Mirza, Mehdi and Xiao, Da and Courville, Aaron and Bengio, Yoshua},
  journal={arXiv preprint arXiv:1312.6211},
  year={2013}
}

@inproceedings{zenke2017continual,
  title={Continual learning through synaptic intelligence},
  author={Zenke, Friedemann and Poole, Ben and Ganguli, Surya},
  booktitle={International conference on machine learning},
  pages={3987--3995},
  year={2017},
  organization={PMLR}
}

@misc{schedule,
      title={Learn the Time to Learn: Replay Scheduling in Continual Learning}, 
      author={Marcus Klasson and Hedvig Kjellström and Cheng Zhang},
      year={2023},
      eprint={2209.08660},
      archivePrefix={arXiv},
      primaryClass={cs.LG}
}

@article{khetarpal2022towards,
  title={Towards continual reinforcement learning: A review and perspectives},
  author={Khetarpal, Khimya and Riemer, Matthew and Rish, Irina and Precup, Doina},
  journal={Journal of Artificial Intelligence Research},
  volume={75},
  pages={1401--1476},
  year={2022}
}

@book{ring1994continual,
  title={Continual learning in reinforcement environments},
  author={Ring, Mark Bishop},
  year={1994},
  publisher={The University of Texas at Austin}
}

@article{parisi2019continual,
  title={Continual lifelong learning with neural networks: A review},
  author={Parisi, German I and Kemker, Ronald and Part, Jose L and Kanan, Christopher and Wermter, Stefan},
  journal={Neural networks},
  volume={113},
  pages={54--71},
  year={2019},
  publisher={Elsevier}
}

@article{abraham2005memory,
  title={Memory retention--the synaptic stability versus plasticity dilemma},
  author={Abraham, Wickliffe C and Robins, Anthony},
  journal={Trends in neurosciences},
  volume={28},
  number={2},
  pages={73--78},
  year={2005},
  publisher={Elsevier}
}

@article{modellingtask,
  title={Continual learning by modeling intra-class variation},
  author={Yu, Longhui and Hu, Tianyang and Hong, Lanqing and Liu, Zhen and Weller, Adrian and Liu, Weiyang},
  journal={arXiv preprint arXiv:2210.05398},
  year={2022}
}

@book{chen2018lifelong,
  title={Lifelong machine learning},
  author={Chen, Zhiyuan and Liu, Bing},
  volume={1},
  year={2018},
  publisher={Springer}
}

@incollection{mccloskey1989catastrophic,
  title={Catastrophic interference in connectionist networks: The sequential learning problem},
  author={McCloskey, Michael and Cohen, Neal J},
  booktitle={Psychology of learning and motivation},
  volume={24},
  pages={109--165},
  year={1989},
  publisher={Elsevier}
}

@misc{loo2020generalized,
      title={Generalized Variational Continual Learning}, 
      author={Noel Loo and Siddharth Swaroop and Richard E. Turner},
      year={2020},
      eprint={2011.12328},
      archivePrefix={arXiv},
      primaryClass={cs.LG}
}

\newpage
\appendix
\label{appendix}

\section{Appendix}

\subsection{Observation of task differences}

\begin{figure}[H]
  \centering
  \includegraphics[scale=0.5]{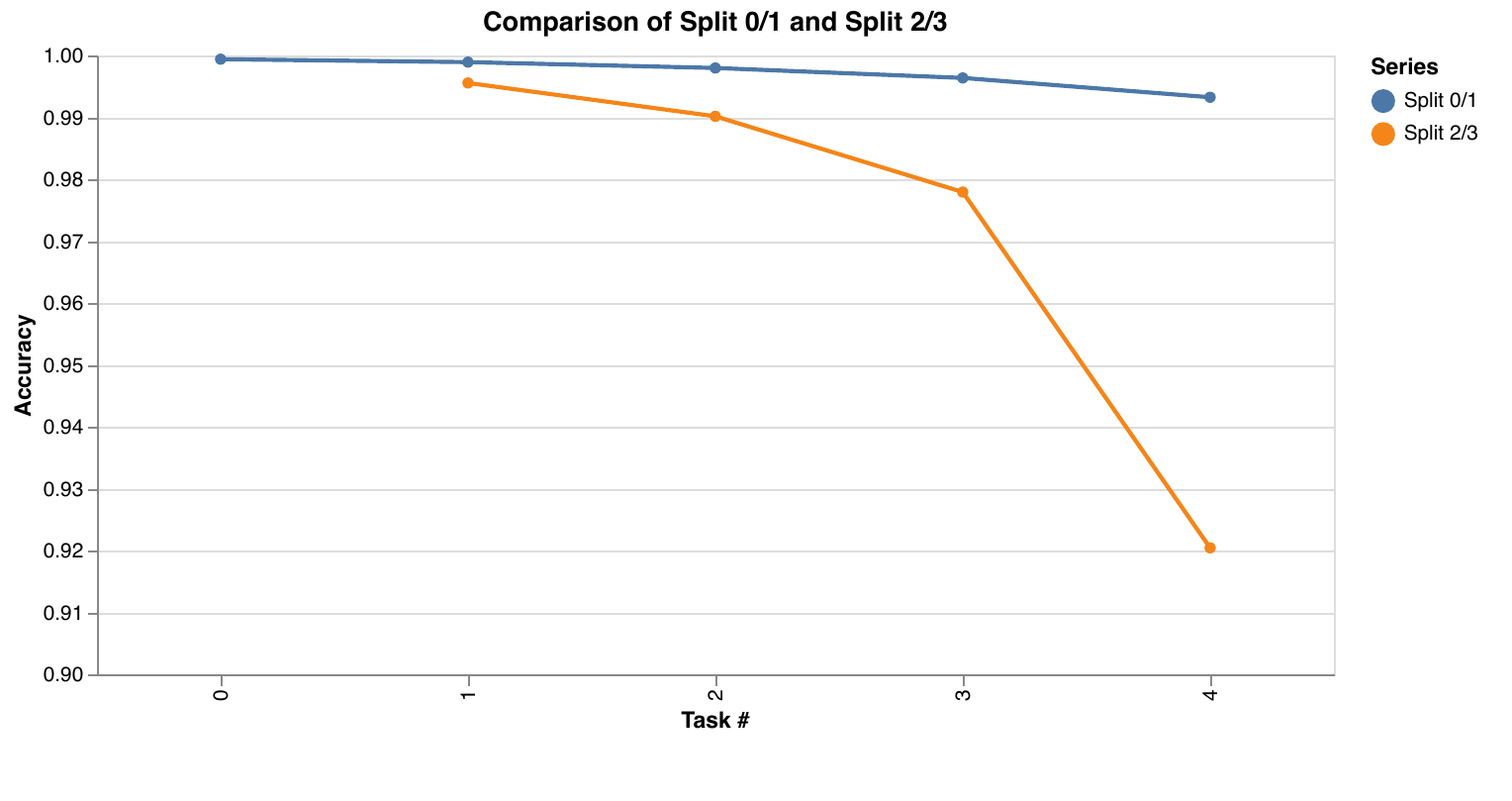}
  \caption{Comparison between tasks 0/1 and 2/3 in the split MNIST experiment. Split 2/3 appears to be a harder task than split 0/1 as its prediction accuracy drops quickly.}
  
\label{mot}
\end{figure}

The study is inspired by the observation of the split-MNIST experiment in the original VCL work, where the task distinguishing 0/1 and 2/3 have distinctive learning trajectories as more tasks get added. Figure \ref{mot} shows that as more tasks are learned, task 0/1 remains relatively high accuracy but task 2/3 is being forgotten quickly.

\subsection{Additional experimental details}

We describe further details relating to the experiments in this section. The code implementation can be found in our \href{https://github.com/lukeyf/auto_vcl}{GitHub repository.}

To generate the difficulty metric, we repeated the mock training with a dataset of size 1000 and a batch size of 256 for one epoch only. We repeated this step 10 times to obtain an average metric for incoming task difficulties. We used an Adam optimizer with a learning rate of 0.001. For the similarity metric, we evaluated the performance of the incoming task once using the existing model without any training.

During the training process, each task was exposed to all training data and trained over 10 epochs. We used a batch size of 256 and an Adam optimizer with a learning rate of 0.001.

For the CIFAR-10 dataset, we preprocessed it by converting it to grayscale and resizing the images.

\subsection{Compute resources}

The experiments were carried on a 2022 MacBook Air with a Apple M2 chip without MPS acceleration. The estimated time to replicate all experiments is under 3 CPU hours. We also noted that applying CUDA GPUs does not necessarily speed up the training. It might be due to that the network is relatively simple and the bottleneck becomes CPU-GPU data transmission.

\subsection{Experiment statistical significance}

In this section, we report the standard error of the mean. Table \ref{split}, \ref{mnist}, and \ref{mixed} shows the numerical accuracy traces of an experimental run with averaged standard error.

\begin{table}[htbp]
\caption{Summary of Average Accuracy and Standard Error across tasks in the Split MNIST experiment with custom targets.}
\label{split}
\centering
\begin{tabular}{lllll}
\toprule
\# of Tasks & GVCL $(\beta=0.01)$ & GVCL $(\beta=1)$ & GVCL $(\beta=100)$ & AutoVCL \\
\midrule
1 & $99.92 \%\pm 0.00 \%$ & $99.92 \%\pm 0.00 \%$ & $99.92 \%\pm 0.00 \%$ & $99.93 \%\pm 0.00 \%$ \\
2 & $99.45 \%\pm 0.21 \%$ & $99.67 \%\pm 0.08 \%$ & $99.13 \%\pm 0.58 \%$ & $99.73 \%\pm 0.09 \%$ \\
3 & $97.60 \%\pm 1.16 \%$ & $98.97 \%\pm 0.29 \%$ & $97.94 \%\pm 0.92 \%$ & $99.05 \%\pm 0.20 \%$ \\
4 & $96.31 \%\pm 1.42 \%$ & $97.47 \%\pm 1.26 \%$ & $97.23 \%\pm 1.14 \%$ & $98.75 \%\pm 0.30 \%$ \\
5 & $92.91 \%\pm 3.27 \%$ & $95.17 \%\pm 2.99 \%$& $96.73 \%\pm 1.15 \%$ & $97.22 \%\pm 1.06 \%$ \\
\bottomrule
\end{tabular}
\end{table}

\begin{table}[htbp]
\caption{Summary of Average Accuracy and Standard Error across tasks in the Permuted MNIST experiment.}
\label{mnist}
\centering
\begin{tabular}{lllll}
\toprule
\# of Tasks & GVCL $(\beta=0.01)$ & GVCL $(\beta=1)$ & GVCL $(\beta=100)$ & AutoVCL \\
\midrule
1 & $97.79 \%\pm 0.00 \%$ & $97.79 \%\pm 0.00 \%$ & $97.79 \%\pm 0.00 \%$ & $97.71 \%\pm 0.00 \%$ \\
2 & $95.86 \%\pm 1.28 \%$ & $96.84 \%\pm 0.46 \%$ & $85.49 \%\pm 8.67 \%$ & $96.69 \%\pm 0.51 \%$ \\
3 & $89.93 \%\pm 4.84 \%$ & $96.12 \%\pm 0.25 \%$ & $81.04 \%\pm 6.83 \%$& $95.91 \%\pm 0.36 \%$ \\
4 & $83.55 \%\pm 7.88 \%$ & $95.51 \%\pm 0.23 \%$ & $78.27 \%\pm 5.62 \%$ & $95.47 \%\pm 0.24 \%$ \\
5 & $78.60 \%\pm 8.73 \%$ & $94.87 \%\pm 0.34 \%$& $77.07 \%\pm 4.64 \%$ & $94.81 \%\pm 0.22 \%$ \\
6 & $73.00 \%\pm 9.32 \%$ & $93.85 \%\pm 0.61 \%$ & $76.00 \%\pm 3.98 \%$ & $93.99 \%\pm 0.30 \%$ \\
7 & $68.03 \%\pm 8.71 \%$ & $92.94 \%\pm 0.75 \%$ & $74.91 \%\pm 3.54 \%$ & $93.13 \%\pm 0.44 \%$\\
8 & $63.91 \%\pm 8.66 \%$ & $92.15 \%\pm 0.87 \%$ & $74.22 \%\pm 3.16 \%$ & $92.43 \%\pm 0.54 \%$ \\
9 & $60.43 \%\pm 8.52 \%$ & $91.01 \%\pm 1.12 \%$ & $73.39 \%\pm 2.89 \%$ & $91.75 \%\pm 0.62 \%$ \\
10 & $58.03 \%\pm 8.17 \%$ & $89.86 \%\pm 1.30 \%$& $72.69 \%\pm 2.65 \%$ & $90.94 \%\pm 0.72 \%$ \\
\bottomrule
\end{tabular}
\end{table}

\begin{table}[H]
\caption{Summary of Average Accuracy and Standard Error across tasks in the mixed experiment with MNIST and CIFAR-10.}
\label{mixed}
\centering
\begin{tabular}{lllll}
\toprule
\# of Tasks & GVCL $(\beta=0.01)$ & GVCL $(\beta=1)$ & GVCL $(\beta=100)$ & AutoVCL \\
\midrule
1 & $99.94 \%\pm 0.00 \%$ & $99.94 \%\pm 0.00 \%$ & $99.94 \%\pm 0.00 \%$ & $99.93 \%\pm 0.00 \%$ \\
2 & $93.37 \%\pm 4.59 \%$ & $92.92 \%\pm 4.94 \%$ & $87.98 \%\pm 8.47 \%$ & $93.52 \%\pm 4.52 \%$ \\
3 & $93.51 \%\pm 5.05 \%$ & $93.85 \%\pm 4.82 \%$ & $91.03 \%\pm 6.42 \%$& $95.30 \%\pm 3.41 \%$ \\
4 & $85.94 \%\pm 6.46 \%$ & $86.35 \%\pm 6.31 \%$ & $84.15 \%\pm 7.57 \%$ & $86.45 \%\pm 6.13 \%$ \\
5 & $86.07 \%\pm 6.17 \%$ & $87.10 \%\pm 6.34 \%$& $86.86 \%\pm 6.69 \%$ & $88.89 \%\pm 5.44 \%$\\
6 & $82.75 \%\pm 5.93 \%$ & $82.03 \%\pm 5.95 \%$ & $82.87 \%\pm 6.46 \%$ & $82.80 \%\pm 4.80 \%$\\
7 & $75.90 \%\pm 5.83 \%$ & $83.46 \%\pm 5.81 \%$ & $85.02 \%\pm 6.01 \%$ & $85.47 \%\pm 4.80 \%$\\
8 & $78.95 \%\pm 5.60 \%$ & $82.13 \%\pm 5.42 \%$ & $83.44 \%\pm 5.47 \%$ & $84.39 \%\pm 4.68 \%$\\
9 & $78.55 \%\pm 5.51 \%$ & $82.75 \%\pm 5.02 \%$ & $84.46 \%\pm 5.05 \%$ & $85.52 \%\pm 4.40 \%$\\
10 & $79.73 \%\pm 4.29 \%$ & $81.85 \%\pm 4.10 \%$& $83.73 \%\pm 4.59 \%$ & $84.94 \%\pm 3.91 \%$ \\
\bottomrule
\end{tabular}
\end{table}

Note that due to the randomness of the training, the traces may appear slightly different for each run especially for models with small $\beta$s (e.g. $\beta=0.01.)$ However, the relative ranking of performances between the four tested models should be consistent.
\newpage
\newpage

\end{document}